# Factorized Asymptotic Bayesian Hidden Markov Models

**Ryohei Fujimaki**                                                                                              RFUJIMAKI@SV.NEC-LABS.COM
NEC Laboratories America, 10080 North Wolfe Road SW3-350, Cupertino, CA 95014 USA

**Kohei Hayashi**                                                                                                 HAYASHI.KOHEI@GMAIL.COM
University of Tokyo

## Abstract

This paper addresses the issue of model selection for hidden Markov models (HMMs). We generalize factorized asymptotic Bayesian inference (FAB), which has been recently developed for model selection on independent hidden variables (i.e., mixture models), for time-dependent hidden variables. As with FAB in mixture models, FAB for HMMs is derived as an iterative lower bound maximization algorithm of a factorized information criterion (FIC). It inherits, from FAB for mixture models, several desirable properties for learning HMMs, such as asymptotic consistency of FIC with marginal log-likelihood, a shrinkage effect for hidden state selection, monotonic increase of the lower FIC bound through the iterative optimization. Further, it does not have a tunable hyper-parameter, and thus its model selection process can be fully automated. Experimental results shows that FAB outperforms states-of-the-art variational Bayesian HMM and non-parametric Bayesian HMM in terms of model selection accuracy and computational efficiency.

## 1. Introduction

An important challenge in learning hidden Markov models (HMMs) is model selection of the number of hidden states. A well-known difficulty is non-regularity in their maximum likelihood (ML) estimators, under which classical information criteria such as Bayes information criterion (BIC) (Schwarz, 1978) lose their theoretical justifications[1].

---
[1] Roughly speaking, Fisher information matrices around the ML estimators are singular, and thus an asymptotic second order approximation is not applicable.

Bayesian inference provides a natural and sophisticated way to address the issue by selecting the model which maximizes marginal log-likelihood (equivalent to the logarithm of the model posterior probability with an uniform model prior). Markov chain Monte Carlo methods (MCMCs) (Robert et al., 2000) and variational Bayesian inference (VB) (MacKay, 1997; Beal, 2003) approximate computationally and analytically intractable marginal log-likelihoods, using respectively, sampling and variational approximation techniques. The former has an advantage over the latter in approximation accuracy but has a disadvantage in computational efficiency, and thus the choice of appropriate inference algorithm has been decided on the basis of a trade-off between accuracy and computational efficiency. In terms of modeling, infinite HMMs (iHMMs) employ a hierarchical Dirichlet process prior in order to express an infinite number of hidden states (Beal et al., 2002). In them, the number of components is determined on the basis of mild prior knowledge expressed by a few hyper-parameters. The state-of-the-art inference of iHMMs proposed by van Gael et al. (2008) uses a beam sampling technique which is more efficient than well-known Gibbs sampling techniques (Beal et al., 2002). Although the beam sampling technique considerably reduces the computational cost of MCMC inference, it is still higher than that of HMMs using variational non-parametric Bayesian inference (VBHMMs), while acceleration of iHMMs has been discussed from the viewpoints of parallelization (Bratieres et al., 2010). In addition, iHMMs have a few hyper-parameters which mildly control the number of hidden states, and determination of them requires further computational costs.

Fujimaki and Morinaga (2012) have recently proposed a new Bayesian approximation inference method for mixture models. They use the terms *factorized information criterion* (FIC) and *factorized asymptotic Bayesian inference* (FAB). FIC is an asymptotically-consistent approximation of marginal log-likelihood using the "factorized" Laplace method, and FAB is its asymptotically-consistent lower bound maximiza-tion algorithm. FAB has been reported to outperform state-of-the-art variational Bayesian method (Fujimaki & Morinaga, 2012). Hereinafter, we denote FIC and FAB for mixture models as $\text{FIC}_{mm}$ and $\text{FAB}_{mm}$, and those for HMMs as $\text{FIC}_{hmm}$ and $\text{FAB}_{hmm}$.

This paper generalizes FIC and FAB for learning HMMs which contain time dependent hidden variables, in contrast to $\text{FAB}_{mm}$, which requires mutual independencies among hidden variables. A key observation is that a "factorized" Laplace method is applicable by decomposing, using the Markov property of hidden states, the complete joint distribution in a specific form of a variational lower bound. $\text{FIC}_{hmm}$ can then be derived as an asymptotic approximation of marginal log-likelihoods of HMMs. The iterative optimization of $\text{FAB}_{hmm}$ can be seen as a natural generalization of the expectation-maximization (EM) algorithm (Dempster et al., 1977), and, interestingly, unique regularizers appear as exponentiated update terms in our FAB forward-backward algorithm. Similar to $\text{FAB}_{mm}$, $\text{FAB}_{hmm}$ has several desirable properties for learning HMMs, such as asymptotic consistency of $\text{FIC}_{hmm}$ with marginal log-likelihood, a shrinkage effect for hidden state selection, and monotonic increase of the lower bound of $\text{FIC}_{hmm}$ through the iterative optimization. An advantage over iHMM is that it has no hyper-parameter, and thus its model selection process can be fully automated though we understand that prior knowledge injection can also be an advantage because we can control models. Further, our experimental results show that model selection accuracy of FAB is competitive to or even better than iHMMs, with significantly-lower computational costs.

## 2. Preliminaries

Let $X = X^1, \ldots, X^T$ and $Z = Z^1, \ldots, Z^T$ be respective sequences of observed and hidden random variables. $Z^t = (Z_1^t, \ldots, Z_K^t)$ is an indicator vector, and $Z_k^t = 1$ if $X^t$ is generated from the $k$-th hidden state, and $Z_k^t = 0$ otherwise. We denote the number of hidden states as $K$. Let us assume that we observe independent $N$ sequences[2] and denote them as $\boldsymbol{x}^N = \boldsymbol{x}_1, \ldots, \boldsymbol{x}_N$. The $n$-th sequence is denoted as $\boldsymbol{x}_n = \boldsymbol{x}_n^1, \ldots, \boldsymbol{x}_n^{T_n}$, where $T_n$ is the length of the $n$-th sequence. We further denote the sequence of latent variables corresponding to $\boldsymbol{x}^N$ and $\boldsymbol{x}_n$ as $\boldsymbol{z}^N = \boldsymbol{z}_1, \ldots, \boldsymbol{z}_N$ and $\boldsymbol{z}_n = \boldsymbol{z}_n^1, \ldots, \boldsymbol{z}_n^{T_n}$, respectively.

An HMM is described as $p(X|\theta) = \sum_Z p(X, Z|\theta) = \sum_Z p(Z^1|\boldsymbol{\alpha}) p(X^1|Z^1, \boldsymbol{\phi}) \prod_{t=2}^T p(Z^t|Z^{t-1}, \boldsymbol{\beta}) p(X^t|Z^t, \boldsymbol{\phi})$

---

[2]Technically, one sequence alone is insufficient for our asymptotic approximation of an initial state probability.

where $\boldsymbol{\theta} = (\boldsymbol{\alpha}, \boldsymbol{\beta}, \boldsymbol{\phi})$. $p(Z^1|\boldsymbol{\alpha})$, $p(Z^t|Z^{t-1}, \boldsymbol{\beta})$ and $p(X^t|Z^t, \boldsymbol{\phi})$ are, respectively, referred to as an initial probability, a transition probability, and an emission probability, and they are, respectively, described as $p(Z^1|\boldsymbol{\alpha}) = \prod_{k=1}^K \alpha_k^{Z_k^1}$, $p(X^t|Z^t, \boldsymbol{\phi}) = \prod_{k=1}^K p(X^t|\boldsymbol{\phi}_k)^{Z_k^t}$, and $p(Z^t|Z^{t-1}, \boldsymbol{\beta}) = \prod_{k=1}^K p_k(Z^t|\boldsymbol{\beta}_k)^{Z_k^{t-1}} = \prod_{j=1}^K \prod_{k=1}^K \beta_{kj}^{Z_j^t Z_k^{t-1}}$, where we define $p_k(Z^t|\boldsymbol{\beta}_k)$ as $p_k(Z^t|\boldsymbol{\beta}_k) \equiv p(Z^t|Z_k^{t-1} = 1, \boldsymbol{\beta}_k)$. $\boldsymbol{\alpha} = (\alpha_1, \ldots, \alpha_K)$, $\boldsymbol{\beta} = (\boldsymbol{\beta}_1, \ldots, \boldsymbol{\beta}_K)$, and $\boldsymbol{\phi} = (\boldsymbol{\phi}_1, \ldots, \boldsymbol{\phi}_K)$ are respective parameters ($\boldsymbol{\beta}_k = (\beta_{k1}, \ldots, \beta_{kK})$). The parameters $\boldsymbol{\alpha}$ and $\boldsymbol{\beta}$ satisfy $\sum_{k=1}^K \alpha_k = 1$ and $\sum_{j=1}^K \beta_{kj} = 1$, respectively. A standard parameter inference follows the EM algorithm with a specific expectation step known as either the forward-backward algorithm (Rabiner, 1989) or the Baum-Welch algorithm (Baum, 1972).

Let us make a few mild assumptions: **A1** the transition matrix $\boldsymbol{\beta}$ is row-independent (i.e., $\boldsymbol{\beta}_{k_1}$ and $\boldsymbol{\beta}_{k_2}$ are mutually independent), **A2** $p(X, Z|\boldsymbol{\theta})$ is bounded (does not diverge to infinity), and **A3** $p(X|\boldsymbol{\phi}_k)$ satisfies the regularity condition. **A1** and **A2** are usual assumptions in HMMs. **A3** is much milder than a regularity assumption on $p(X|\boldsymbol{\theta})$, and many HMMs (e.g., a HMM with categorical observations, a HMM with Gaussian observations) satisfy this assumption.

Let us denote a model of $p(X|\theta)$ as $M$. We allow $K$ emission probabilities to be different from one another in their model representations (e.g., for logistic regression emissions, different hidden states can have feature configurations with different complexities). This can be seen as an HMM-extension of the so-called "heterogeneous mixture models" (Fujimaki et al., 2011). In order to distinguish different model representations, we denote a model of $\boldsymbol{\phi}_k$ as $S_k$. That is, our model $M$ is specified by $K$ and emission models $S_k$, i.e., $M = (K, S_1, \ldots, S_K)$. Although parameter representations depend on the corresponding models, we hereinafter omit them for notational simplicity.

## 3. FIC for HMMs

$\text{FIC}_{hmm}$ considers the following lower bound of marginal log-likelihood as:

$$\log p(\boldsymbol{x}^N|M) \geq \sum_{\boldsymbol{z}^N} q(\boldsymbol{z}^N) \log\left(\frac{p(\boldsymbol{x}^N, \boldsymbol{z}^N|M)}{q(\boldsymbol{z}^N)}\right) \quad (1)$$

In contrast to the standard VB lower bound, $\sum_{\boldsymbol{z}^N} \int q(\boldsymbol{z}^N) q(\boldsymbol{\theta}) \log\left(p(\boldsymbol{x}^N, \boldsymbol{z}^N, \boldsymbol{\theta}|M)/q(\boldsymbol{z}^N)q(\boldsymbol{\theta})\right) d\boldsymbol{\theta}$, which makes a mutual independence assumption between $\boldsymbol{z}^N$ and $\boldsymbol{\theta}$ on variational distributions, this lower bound does not. Although (1) is also used in $\text{FIC}_{mm}$ and collapse VBs (Teh et al., 2006; Kurihara et al., 2007), both of them require mutual independences among hidden variables, and thus they are not directly applicable for models having time-dependent hidden variables (e.g., HMMs).

On the basis of the Markov property of hidden states, **A1**, and **A2**, the numerator of (1) is factorized as follows:

$$p(\boldsymbol{x}^N, \boldsymbol{z}^N | M) = \int \prod_{n=1}^{N} \Big\{ p(\boldsymbol{z}_n^1 | \boldsymbol{\alpha}) \prod_{k=1}^{K} \prod_{t=2}^{T_n} p_k(\boldsymbol{z}_n^t | \boldsymbol{\beta}_k)^{z_{nk}^{t-1}}$$

$$\times \prod_{k=1}^{K} \prod_{t=1}^{T_n} p(\boldsymbol{x}_n^t | \boldsymbol{\phi}_k)^{z_{nk}^t} \Big\} p(\boldsymbol{\theta} | M) d\boldsymbol{\theta}, \quad (2)$$

where $p(\boldsymbol{\theta}|M) = p(\boldsymbol{\alpha}|M) \prod_{k=1}^{K} p(\boldsymbol{\beta}_k|M) p(\boldsymbol{\phi}_k|M)$ is a parameter prior, and we assume that the priors' dimensionalities are asymptotically small. Each one of the factorized distributions, $p(\boldsymbol{z}_n^1|\boldsymbol{\alpha})$, $p_k(\boldsymbol{z}_n^t|\boldsymbol{\beta}_k)$ and $p(\boldsymbol{x}_n^t|\boldsymbol{\phi}_k)$, is a regular model (**A3**) to which the Laplace method is applicable. Then, as with FIC$_{mm}$, the "factorized" Laplace method around the ML estimator[3] $\bar{\boldsymbol{\theta}}$ of $p(\boldsymbol{x}^N, \boldsymbol{z}^N|\boldsymbol{\theta})$ gives us a second order approximation of complete likelihood as follows:

$$\log p(\boldsymbol{x}^N, \boldsymbol{z}^N|\boldsymbol{\theta}) \approx \log p(\boldsymbol{x}^N, \boldsymbol{z}^N|\bar{\boldsymbol{\theta}}) - \frac{N}{2}[\bar{\mathcal{F}}_{\boldsymbol{\alpha}}]$$

$$- \sum_{k=1}^{K} \Big( \frac{\sum_{n,t=1}^{N,T_n-1} z_{nk}^t}{2} [\bar{\mathcal{F}}_{\boldsymbol{\beta}_k}] - \frac{\sum_{n,t=1}^{N,T_n} z_{nk}^t}{2} [\bar{\mathcal{F}}_{\boldsymbol{\phi}_k}] \Big). \quad (3)$$

The notation $[\bar{\mathcal{F}}_\bullet]$ represents a (centered) quadratic form with respect to $\bullet$, e.g., $[\bar{\mathcal{F}}_{\boldsymbol{\alpha}}] = (\boldsymbol{\alpha} - \bar{\boldsymbol{\alpha}})^T \bar{\mathcal{F}}_{\boldsymbol{\alpha}} (\boldsymbol{\alpha} - \bar{\boldsymbol{\alpha}})$. $\bar{\mathcal{F}}_{\boldsymbol{\alpha}}$, $\bar{\mathcal{F}}_{\boldsymbol{\beta}_k}$, and $\bar{\mathcal{F}}_{\boldsymbol{\phi}_k}$ are Fisher information matrices approximated by (complete) data instances as follows:

$$\bar{\mathcal{F}}_{\boldsymbol{\alpha}} = \frac{-1}{N} \nabla_{\boldsymbol{\alpha}}^2 \sum_{n=1}^{N} \log p(\boldsymbol{z}_n^1|\boldsymbol{\alpha}) \Big|_{\boldsymbol{\alpha}=\bar{\boldsymbol{\alpha}}} \quad (4)$$

$$\bar{\mathcal{F}}_{\boldsymbol{\beta}_k} = \frac{-1}{\sum_{n,t=1}^{N,T_n-1} z_{nk}^t} \nabla_{\boldsymbol{\beta}_k}^2 \sum_{n,t=1}^{N,T_n-1} z_{nk}^t \log p_k(\boldsymbol{z}_n^{t+1}|\boldsymbol{\beta}_k) \Big|_{\boldsymbol{\beta}_k=\bar{\boldsymbol{\beta}}_k}$$

$$\bar{\mathcal{F}}_{\boldsymbol{\phi}_k} = \frac{-1}{\sum_{n,t=1}^{N,T_n} z_{nk}^t} \nabla_{\boldsymbol{\phi}_k}^2 \sum_{n,t=1}^{N,T_n} z_{nk}^t \log p(\boldsymbol{x}_n^t|\boldsymbol{\phi}_k) \Big|_{\boldsymbol{\phi}_k=\bar{\boldsymbol{\phi}}_k}.$$

It is easy to show that they respectively converge to Fisher information matrices of $p(Z^1|\boldsymbol{\alpha})$, $p_k(Z^t|\boldsymbol{\beta}_k)$, and $p(X|\boldsymbol{\phi}_k)$ with $N \to \infty$, and it guarantees their determinants to be $\mathcal{O}(1)$. The complete marginal likelihood can then be asymptotically approximated as:

$$p(\boldsymbol{x}^N, \boldsymbol{z}^N|M) \approx p(\boldsymbol{x}^N, \boldsymbol{z}^N|\bar{\boldsymbol{\theta}}) \frac{(2\pi)^{\mathcal{D}_{\boldsymbol{\alpha}}/2}}{N^{\mathcal{D}_{\boldsymbol{\alpha}}/2} |\bar{\mathcal{F}}_{\boldsymbol{\alpha}}|^{1/2}}$$

$$\times \prod_{k=1}^{K} \frac{(2\pi)^{\mathcal{D}_{\boldsymbol{\beta}_k}/2}}{(\sum_{n,t=1}^{N,T_n-1} z_{nk}^t)^{\mathcal{D}_{\boldsymbol{\beta}_k}/2} |\bar{\mathcal{F}}_{\boldsymbol{\beta}_k}|^{1/2}}$$

$$\times \prod_{k=1}^{K} \frac{(2\pi)^{\mathcal{D}_{\boldsymbol{\phi}_k}/2}}{(\sum_{n,t=1}^{N,T_n} z_{nk}^t)^{\mathcal{D}_{\boldsymbol{\phi}_k}/2} |\bar{\mathcal{F}}_{\boldsymbol{\phi}_k}|^{1/2}}. \quad (5)$$

Here, $\mathcal{D}_\bullet$ is the dimensionality of $\bullet$ ($\mathcal{D}_{\boldsymbol{\alpha}} = K-1$ and $\mathcal{D}_{\boldsymbol{\beta}_k} = K-1$).

By substituting (5) into (1) and ignoring asymptotically small terms, we obtain an asymptotic approximation of $\log p(\boldsymbol{x}^N|M)$ as follows:

$$FIC(\boldsymbol{x}^N, M) \equiv \max_q \{\mathcal{J}(q, \bar{\boldsymbol{\theta}}, \boldsymbol{x}^N)\} \quad (6)$$

$$\mathcal{J}(q, \bar{\boldsymbol{\theta}}, \boldsymbol{x}^N) = \sum_{\boldsymbol{z}^N} q(\boldsymbol{z}^N) \Big( \log p(\boldsymbol{x}^N, \boldsymbol{z}^N|\bar{\boldsymbol{\theta}})$$

$$- \frac{\mathcal{D}_{\boldsymbol{\alpha}}}{2} \log N - \sum_{k=1}^{K} \frac{\mathcal{D}_{\boldsymbol{\beta}_k}}{2} \log\Big( \sum_{n,t=1}^{N,T_n-1} z_{nk}^t \Big)$$

$$- \sum_{k=1}^{K} \frac{\mathcal{D}_{\boldsymbol{\phi}_k}}{2} \log\Big( \sum_{n,t=1}^{N,T_n} z_{nk}^t \Big) - \log q(\boldsymbol{z}^N) \Big). \quad (7)$$

We here have two regularization terms dependent on hidden states, $\mathcal{D}_{\boldsymbol{\beta}_k} \log(\sum_{n,t=1}^{N,T_n-1} z_{nk}^t)/2$ and $\mathcal{D}_{\boldsymbol{\phi}_k} \log(\sum_{n,t=1}^{N,T_n} z_{nk}^t)/2$. The latter stems from asymptotic approximation of the emission (observation) probability, and appears in FIC$_{mm}$. Contrastingly, the former is an unique regularization term in FIC$_{hmm}$. Notably, these two regularizers contain dependencies between parameters ($\boldsymbol{\beta}_k$ and $\boldsymbol{\phi}_k$) and hidden states, which the variational lower bound of VB methods usually ignore on their variational distributions (Beal, 2003). These regularizers will be discussed in more detail in sub-sections 4.4 and 4.5

The following theorem justifies FIC$_{hmm}$ as an approximation of marginal log-likelihood:

**Theorem 1** $FIC(\boldsymbol{x}^N, M)$ *is asymptotically consistent with* $\log p(\boldsymbol{x}^N|M)$.

A rough sketch of the theorem can be described as follows. Because of the regularity condition **A3**, individual Laplace approximations have asymptotic consistency, and thus their product (5) is also consistent with $p(\boldsymbol{x}^N, \boldsymbol{z}^N|M)$. Then, since there is a $q$ which satisfies the equality in (1), the theorem holds.

---

[3] As an alternative, we could employ the maximum-a-posteriori estimator of $p(\boldsymbol{x}^N, \boldsymbol{z}^N|\boldsymbol{\theta})p(\boldsymbol{\theta})$. This would not make a significant difference with respect to discussion in this paper.

## 4. FAB for HMMs

### 4.1. FAB Lower Bound

Since $\bar{\boldsymbol{\theta}}$ is defined on both $\boldsymbol{x}^N$ and $\boldsymbol{z}^N$ and thus is not available in practice, we cannot evaluate $\text{FIC}_{hmm}$ itself. Instead, FAB maximizes the lower bound of $\text{FIC}_{hmm}$. As is similarly done in (Fujimaki & Morinaga, 2012), we employ two inequalities to derive the lower bound. First, on the basis of the definition of the ML estimator, $\log p(\boldsymbol{x}^N, \boldsymbol{z}^N | \bar{\boldsymbol{\theta}}) \geq \log p(\boldsymbol{x}^N, \boldsymbol{z}^N | \boldsymbol{\theta})$ holds. Second, on the basis of the concavity of the logarithm function, $\log(\sum_{n,t=1}^{N,T_n} z_{nk}^t) \geq \mathcal{L}(\sum_{n,t=1}^{N,T_n} z_{nk}^t, \sum_{n,t=1}^{N,T_n} \tilde{q}_{nk}^t)$ holds with an arbitrary $\tilde{q}$ (the same holds for $\log(\sum_{n,t=1}^{N,T_n-1} z_{nk}^t)$), where $\mathcal{L}(a,b) = \log b + (a-b)/b$. The lower bound of (6) is then derived as follows:

$$FIC(\boldsymbol{x}^N, M) \geq \mathcal{G}(q, \tilde{q}, \boldsymbol{x}^N, \boldsymbol{\theta}) \tag{8}$$

$$= \sum_{n=1}^{N} \sum_{\boldsymbol{z}_n} q(\boldsymbol{z}_n) \Big[ \log p(\boldsymbol{x}_n, \boldsymbol{z}_n | \boldsymbol{\theta}) + \sum_{k,t=1}^{K,T_n} z_{nk}^t \log \delta_k^t$$

$$- \log q(\boldsymbol{z}^N) \Big] + \sum_{n,t=1}^{N,T_n} \log \Delta^t - \frac{\mathcal{D}_{\boldsymbol{\alpha}}}{2} \log N - \sum_{k=1}^{K} \Big( \frac{\mathcal{D}_{\boldsymbol{\beta}_k}}{2} \times$$

$$(\log(\sum_{n,t}^{N,T_n-1} \tilde{q}(z_{nk}^t)) - 1) + \frac{\mathcal{D}_{\boldsymbol{\phi}_k}}{2} (\log(\sum_{n,t}^{N,T_n} \tilde{q}(z_{nk}^t)) - 1) \Big),$$
$$\tag{9}$$

$$\delta_k^t = \begin{cases} \frac{1}{\Delta^t} \exp\Big(-\frac{\mathcal{D}_{\boldsymbol{\beta}_k}}{2(\sum_{n,t=1}^{N,T_n-1} \tilde{q}(z_{nk}^t))} - \frac{\mathcal{D}_{\boldsymbol{\phi}_k}}{2(\sum_{n,t=1}^{N,T_n} \tilde{q}(z_{nk}^t))}\Big) \\ \qquad \text{if } t < T_n \\ \frac{1}{\Delta^t} \exp\Big(-\frac{\mathcal{D}_{\boldsymbol{\phi}_k}}{2(\sum_{n,t=1}^{N,T_n} \tilde{q}(z_{nk}^t))}\Big) \quad \text{if } t = T_n \end{cases},$$
$$\tag{10}$$

where $\Delta^t$ is a normalization constant for $\sum_{k=1}^{K} \delta_k^t = 1$. The underlined part is referred to in (19).

$\text{FAB}_{hmm}$ learns HMMs by solving the following optimization problem (recall that $\boldsymbol{\theta}$ and $q$ are respective functions of $M$):

$$\boldsymbol{M}^*, \boldsymbol{\theta}^*, q^*, \tilde{q}^* = \arg\max_{\boldsymbol{M}, \boldsymbol{\theta}, q, \tilde{q}} \mathcal{G}(q, \tilde{q}, \boldsymbol{x}^N, \boldsymbol{\theta}). \tag{11}$$

Here we have a newly-introduced parameter $\tilde{q}$ which is also optimized in $\text{FAB}_{hmm}$. Let us fix $\boldsymbol{\theta}$ and $q$. By making the differential of (9) with respect to $\tilde{q}(z_{nk}^t)$ zero, we obtain the following optimality conditions:

$$\frac{\mathcal{D}_{\boldsymbol{\phi}_k}}{2} \Big( \frac{1}{\sum_{n,t}^{N,T_n} \tilde{q}(z_{nk}^t)} - \frac{\sum_{n,t}^{N,T_n} q(z_{nk}^t)}{(\sum_{n,t}^{N,T_n} \tilde{q}(z_{nk}^t))^2} \Big) + \tag{12}$$

$$\frac{\mathcal{D}_{\boldsymbol{\beta}_k}}{2} \Big( \frac{1}{\sum_{n,t}^{N,T_n-1} \tilde{q}(z_{nk}^t)} - \frac{\sum_{n,t}^{N,T_n-1} q(z_{nk}^t)}{(\sum_{n,t}^{N,T_n-1} \tilde{q}(z_{nk}^t))^2} \Big) = 0.$$

Clearly, $\tilde{q} = q$ satisfies (12) for arbitrary $\tilde{q}(z_{nk}^t)$. This result will be used in the next subsection.

### 4.2. Iterative Optimization with FAB Forward-Backward Algorithm

Let us first fix $K$ and consider the optimization of (11) with respect to $\boldsymbol{S}, \boldsymbol{\theta}, q, \tilde{q}$, where $\boldsymbol{S} = (S_1, \ldots, S_K)$. Since their simultaneous optimization would be intractable, $\text{FAB}_{hmm}$ works on the basis of iterations of two sub-steps (V-step and M-step). Let the superscription $(i)$ represent the $i$-th iteration.

**V-step (FAB Forward-Backward Algorithm)**
In the $i$-th V-step, we fix $\boldsymbol{\theta} = \boldsymbol{\theta}^{(i-1)}$ and also fix $\tilde{q} = q^{(i-1)}$ on the basis of (12). $\text{FAB}_{hmm}$ then optimizes $q$ by maximization in (9). The terms in (9) dependent on $q$ can be decomposed in terms of sequences, and thus we can independently optimize $q(\boldsymbol{z}_n)$ for individual sequences. Further, the maximization problem for $q(\boldsymbol{z}_n)$ with respect to $\sum_{\boldsymbol{z}_n} q(\boldsymbol{z}_n)[\log p(\boldsymbol{x}_n, \boldsymbol{z}_n | \boldsymbol{\theta}) + \sum_{k,t=1}^{K,T_n} z_{nk}^t \log \delta_k^t - \log q(\boldsymbol{z}_n)]$ has a form similar to the E-step in the EM algorithm for HMMs. In fact, the only difference is $\sum_{k,t=1}^{K,T_n} z_{nk}^t \log \delta_k^t$, which arises from the regularization terms $\mathcal{D}_{\boldsymbol{\beta}_k} \log(\sum_{n,t=1}^{N,T_n-1} z_{nk}^t)/2$ and $\mathcal{D}_{\boldsymbol{\phi}_k} \log(\sum_{n,t=1}^{N,T_n} z_{nk}^t)/2$ in (7).

Notably, the term $z_{nk}^t \log \delta_k^t$ is a product of a hidden variable and a log-probability because $\delta_k^t$ is normalized as is defined in (10). The maximization problem can thus be solved on the basis of the forward-backward algorithm described as follows:

$$f_{nk}^{t(i)} = \begin{cases} \frac{1}{\zeta_n^{t(i)}} \alpha_k^{(i-1)} \tilde{p}(\boldsymbol{x}_n^1 | \boldsymbol{\phi}_k^{(i-1)}) & \text{if } t = 1 \\ \frac{1}{\zeta_n^{t(i)}} \tilde{p}(\boldsymbol{x}_n^t | \boldsymbol{\phi}_k^{(i-1)}) \sum_{j=1}^{K} f_{nj}^{t-1(i)} \beta_{jk}^{(i-1)} \end{cases}$$

$$b_{nk}^{t(i)} = \begin{cases} \frac{1}{\zeta_n^{t+1(i)}} \sum_{j=1}^{K} b_n^{t+1(i)} \tilde{p}(\boldsymbol{x}_n^{t+1} | \boldsymbol{\phi}_j^{(i-1)}) \beta_{kj}^{(i-1)} \\ 1 \quad \text{if } t = T_n \end{cases}$$

$$\tilde{p}(\boldsymbol{x}_n^t | \boldsymbol{\phi}_k^{(i-1)}) = p(\boldsymbol{x}_n^t | \boldsymbol{\phi}_k^{(i-1)}) \delta_k^{t(i-1)}. \tag{13}$$

$\zeta_n^{t(i)}$ is a normalization constant for $\sum_{k=1}^{K} f_{nk}^{t(i)} = 1$. On the basis of $f_{nk}^{t(i)}$ and $b_{nk}^{t(i)}$, the variational distributions are calculated as follows:

$$q^{(i)}(z_{nk}^t) = f_{nk}^{t(i)} b_{nk}^{t(i)} \tag{14}$$

$$q^{(i)}(z_{nj}^{t-1}, z_{nk}^t) = \frac{1}{\zeta_n^{t(i)}} f_{nj}^{t-1(i)} \tilde{p}(\boldsymbol{x}_n^t | \boldsymbol{\phi}_k^{(i-1)}) \beta_{jk}^{(i-1)} b_{nk}^{t(i)}$$

In the above FAB forward-backward algorithm, the only difference from the standard forward-backward algorithm is the exponentiated update term, $\delta_k^{t(i-1)}$. It is interesting that $\text{FAB}_{mm}$ has a similar exponentiated update term in its V-step, and it generates significant differences from standard ML estimation using the EM algorithm. Similarly, $\delta_k^{t(i-1)}$ generates essential differences for learning HMMs between FAB$_{hmm}$ estimation and ML estimation. Roughly speaking, (10) indicates that the smaller and more complex components are likely to become even smaller through the iterations. Further, (13) indicates that the regularization effect of $\delta_k^{t(i-1)}$ propagates forward and backward (e.g., largely-regularized hidden states in $f_{nk}^{t(i)}$ make only small contribution to $f_{nj}^{t+1(i)}$).

**Mstep** Let us fix $q = q^{(i)}$ and $\tilde{q} = q^{(i)}$. FAB$_{hmm}$ then optimizes $\boldsymbol{\theta}$ by maximization in (9). First, we have the following parameter updates for $\boldsymbol{\alpha}$ and $\boldsymbol{\beta}$:

$$\alpha_k^{(i)} \propto \sum_{n=1}^{N} q^{(i)}(z_{nk}^1), \quad \beta_{jk}^{(i)} \propto \sum_{n,t=1}^{N,T_n-1} q^{(i)}(z_{nj}^t, z_{nk}^{t+1}) \quad (15)$$

We then update $S_k$ and $\boldsymbol{\phi}_k$ by solving the following optimization problem:

$$S_k^{(i)}, \boldsymbol{\phi}_k^{(i)} = \arg\max_{S_k, \boldsymbol{\phi}_k} \Big\{ \sum_{n,t=1}^{N,T_n} q^{(i)}(z_{nk}^t)(\log p(\boldsymbol{x}_n^t|\boldsymbol{\phi}_k)$$
$$- \frac{\mathcal{D}_{\boldsymbol{\beta}_k}^{(i)}}{2} \log(\sum_{n,t}^{N,T_n-1} q^{(i)}(z_{nk}^t)) \Big\} \quad (16)$$

It is worth noting that FAB$_{hmm}$ provides a natural way of seeking emission probability types $S_k$ because the M-step can be decomposed into optimization problems for individual hidden states (otherwise, we must take into account an exponential number of hidden state combinations.) With a finite set of emission probability candidates, we first optimize $\boldsymbol{\phi}_k$ for each element of a fixed $S_k$ and then select the optimal one by comparing them. If we employ a standard HMM having single emission type, (16) is reduced to the M-step of the standard EM algorithm for HMMs, i.e., $\boldsymbol{\phi}_k^{(i)} = \arg\max_{\boldsymbol{\phi}_k} \sum_{n,t=1}^{N,T_n} q^{(i)}(z_{nk}^t) \log p(\boldsymbol{x}_n^t|\boldsymbol{\phi}_k)$. Further, FAB$_{hmm}$ is reduced to the standard EM algorithm with $N \to \infty$ (because $\delta_k^t = 1$) and thus can be seen as its natural generalization.

### 4.3. Convergence and Stopping Criterion

Let us denote the lower bound of FIC as follows:

$$FIC_{LB}^{(i)}(\boldsymbol{x}^N, M) \equiv \mathcal{G}(q^{(i)}, \tilde{q}^{(i)} = q^{(i)}, \boldsymbol{x}^N, \boldsymbol{\theta}^{(i)}). \quad (17)$$

Then, as is done for mixture models (Fujimaki & Morinaga, 2012), FAB$_{hmm}$ is guaranteed to monotonically increase $FIC_{LB}^{(i)}(\boldsymbol{x}^N, M)$ through the VM-iterations, which can be summarized in the following theorem.

**Theorem 2** *For the VM iterations of FAB$_{hmm}$, the following inequality is satisfied:*

$$FIC_{LB}^{(i)}(\boldsymbol{x}^N, M) \geq FIC_{LB}^{(i-1)}(\boldsymbol{x}^N, M). \quad (18)$$

We employ $FIC_{LB}^{(i)}(\boldsymbol{x}^N, M) - FIC_{LB}^{(i-1)}(\boldsymbol{x}^N, M) \leq \tau$ as a stopping criterion. One issue is computation of the term $\sum_{\boldsymbol{z}_n} q(\boldsymbol{z}_n) \log q(\boldsymbol{z}_n)$ in (9), which naively requires $\mathcal{O}(K^{T_n})$ computational cost. We can solve this problem using a way similar to that for the variational free energy for VBHMMs (Beal, 2003). In summary, $FIC_{LB}^{(i)}(\boldsymbol{x}^N, M)$ can be computed as follows (we omit the derivation here because of space limitations):

$$FIC_{LB}^{(i)}(\boldsymbol{x}^N, M) = \sum_{n=1}^{N} \sum_{t=1}^{T_n} \log \zeta_n^{t(i)} + \underline{\text{underline of (9)}}. \quad (19)$$

### 4.4. Automatic Hidden State Selection

An interesting property of the asymptotic exponentiated regularizer $\delta_k^t$ (10) is a shrinkage effect for selecting the number of hidden states, which we fixed in the previous section, and thus it provides over-fitting mitigation despite our asymptotic ignoring of priors.

In (13), $f_{nk}^{t(i)}$ with a small $\delta_k^{t(i-1)}$ value is largely regularized without relation to the observation and previous paths. While $b_{nk}^{t(i)}$ does not explicitly have such a regularization effect, one notable fact is that each next path $b_{nj}^{t+1(i)}$ having a large $\delta_j^{t(i+1)}$ value makes only a small contribution to the update of $b_{nk}^{t(i)}$. Then, in (14), the $k$-th hidden state with a small $\delta_k^{t(i-1)}$ value has a small size $\sum_{n,t}^{N,T_n} q^{(i)}(z_{nk}^t)$. Note that, from the definition in (10), the smaller hidden state has the smaller $\delta_k^t$ value. This means such a hidden state gradually becomes smaller through the VM iterations. Similarly, the frequency of transitions between the $j$-th and the $k$-th states having respective small $\delta_j^{t-1}$ and $\delta_k^t$ values becomes smaller because $\sum_{n,t}^{N,T_n-1} q^{(i)}(z_{nj}^t, z_{nk}^{t+1})$ gradually becomes smaller.

On the basis of the above insight, FAB$_{hmm}$ shrinks hidden states using a thresholding operation as follows:

$$q^{(i)}(z_{nk}^t) = 0, q^{(i)}(z_{nk}^t, z_{nj}^{t+1}) = 0 \quad \text{if} \quad \sum_{n,t}^{N,T_n} q^{(i)}(z_{nk}^t) \leq \varepsilon. \quad (20)$$

Starting from an appropriately-large number ($K_{\max}$) of hidden states, FAB$_{hmm}$ iteratively optimizes $\boldsymbol{S}$, $\boldsymbol{\theta}$, and $q$. During the VM steps, the number of hidden states might become smaller due to the shrinkage operation. Then, at a convergence point, we obtain the optimal model $M^* = (K^*, \boldsymbol{S}^*)$, the optimized parameter $\boldsymbol{\theta}^*$, and the variational distribution $q^*$. A similar shrinkage effect has been reported in FAB$_{mm}$ (Fujimaki & Morinaga, 2012), and FAB$_{hmm}$ naturally inherits it for Markov hidden variables.

### 4.5. Discussion: Comparison with VB and BIC

We here compare three approximation inference methods (FAB, VB and BIC) and discuss their differences. FAB approximates marginal log-likelihoods by (19). Let us denote normalization constants for forward-backward algorithms of ML and VB estimations as $\zeta_n^t(ML)$ and $\zeta_n^t(VB)$, respectively. Variational free energy $\mathcal{F}_{VB}$ and BIC[4] can then be respectively computed as follows (see (Beal, 2003) for VB free energy):

$$\mathcal{F}_{VB} = \sum_{n,t=1}^{N,T_n} \log \zeta_n^t(VB) + \int d\boldsymbol{\alpha} q(\boldsymbol{\alpha}) \log \frac{p(\boldsymbol{\alpha})}{q(\boldsymbol{\alpha})} \quad (21)$$

$$+ \int d\boldsymbol{\beta} q(\boldsymbol{\beta}) \log \frac{p(\boldsymbol{\beta})}{q(\boldsymbol{\beta})} + \sum_{k=1}^{K} \int d\boldsymbol{\phi_k} q(\boldsymbol{\phi_k}) \log \frac{p(\boldsymbol{\phi_k})}{q(\boldsymbol{\phi_k})},$$

$$BIC = \sum_{n,t=1}^{N,T_n} \log \zeta_n^t(ML) - \frac{\mathcal{D}}{2} \log \sum_{n=1}^{N} T_n \quad (22)$$

where $\mathcal{D} = \mathcal{D}_\alpha + \sum_{k=1}^{K}(\mathcal{D}_{\beta_k} + \mathcal{D}_{\phi_k})$.

Here (19), (21), and (22) all have closely similar representations, i.e., data fitting term + complexity. The data fitting terms ($\zeta$-related terms) are different in posterior (or variational) distributions of hidden states, on which respective complete log-likelihood functions are marginalized (of course, parameter estimators are also different.) Interesting differences appear in the complexity terms. The complexity term of VB is the Kullback-Leibler divergence between the variational posteriors and priors on parameters. Therefore, VB regularizes parameters (precisely speaking, variational posteriors) to be apart from priors. One disadvantage is that the complexity term is dependent on a choice of priors (and their hyper-parameters), and model selection results will thus be similarly dependent, though we understand that this can also be an advantage because we can control models using priors. On the other hand, FAB does not have manually-tunable parameters in (19). Notably, only the FAB complexity takes into account the distribution on hidden states (i.e., $\tilde{q}^*(\boldsymbol{z}^N) = q^*(\boldsymbol{z}^N)$), and FAB automatically adjusts regularization levels on the basis of sizes ($\sum_{n,t}^{N,T_n} q(z_{nk}^t)$) and the dimensionality of individual hidden states. The regularization term of BIC is stronger than that of FAB (all hidden states are regularized with the scale $\log \sum_{n=1}^{N} T_n$.) Interestingly, the stochastic complexity of HMMs is theoretically proven to be considerably smaller than $\mathcal{D}/2 \log \sum_{n=1}^{N} T_n$ (Yamazaki & Watanabe, 2005). Our result also suggest that a brute-force application of BIC to HMMs overestimates the complexity term.

## 5. Experiments and Discussion

We conducted simulations using artificial data for investigating basic behaviors of HMMs with FAB (FABHMMs), and evaluation using real world e-book data. FABHMM was compared with VBHMM, iHMM, and HMMs with ML estimation and BIC model selection (MLHMMs). We implemented FABHMM and MLHMM by Python, while we used Matlab softwares for iHMM[5] and VBHMM[6].

### 5.1. Simulations with Artificial Data

By following settings similar to those in (van Gael et al., 2008), we conducted evaluations on HMMs with either one-dimensional Gaussian emissions or categorical emissions. The true model had four hidden states, either Gaussian emission probabilities with means $(-4, -1, 2, 3)$ and variances $(0.5, 0.5, 0.5, 0.5)$ or categorical emission probabilities with 8-alphabet, and transition probability described as follows:

$$\begin{pmatrix} 0 & 1 & 1 & 0 \\ 0 & 0 & 1 & 1 \\ 1 & 0 & 0 & 1 \\ 1 & 1 & 0 & 0 \end{pmatrix} /2 \text{ and } \begin{pmatrix} 1 & 0 & 0 & 0 & 0 & 0 & 1 & 1 \\ 1 & 1 & 1 & 0 & 0 & 0 & 0 & 0 \\ 0 & 0 & 1 & 1 & 1 & 0 & 0 & 0 \\ 0 & 0 & 0 & 0 & 1 & 1 & 1 & 0 \end{pmatrix} /3,$$

where the left and right matrices correspond to transition and categorical emission probabilities. We set the initial probability of the first state as one and the rests were zero. We randomly generated a single training sequences with the length of $T \in \{250, 500, 1000, 2000\}$, and a test sequence with the length of $T_{test} = 5000$. For iHMMs, we randomly initialized the hidden state sequences by 10 different states. We set two hyper-priors in iHMMs as (1) $\Gamma(4, 1)$, $\Gamma(4, 1)$ and (2) $\Gamma(1, 0.01)$, $\Gamma(1, 0.01)$, where $\Gamma(a, b)$ denotes the Gamma distribution with the shape parameter $a$ and the scale parameter $b$. The former setting (iHMM1) were used in (van Gael et al., 2008) as a default setting and the latter setting (iHMM2) was less informative. VBHMM and MLHMM were performed with setting $K = 1, \ldots, K_{\max}$ and selected the optimal $K^*$ which minimized the respective free energy and BIC values. For FABHMM, VBHMM and MLHMM, $K_{\max}$ is set to be ten (FAB started from $K_{\max}$ hidden states and automatically searched the optimal value.) We evaluated the estimated number of hidden states (model selection accuracy), predictive log-likelihood (PLL) against the testing set (generalization performance), and training time (computational efficiency). If the training time of each method exceeded 10 minutes, we stopped the training procedure and used the result at the time (iHHM violated this time limit a few times.) The results below are the averages of ten runnings.

---

[4]The (22) representation of BIC does not have theoretical justification for HMMs, as we have noted in Section 1.

[5]iHMM: mloss.org/revision/view/291/
[6]Categorical: www.gatsby.ucl.ac.uk/vbayes, Gaussian: www.robots.ox.ac.uk/~parg

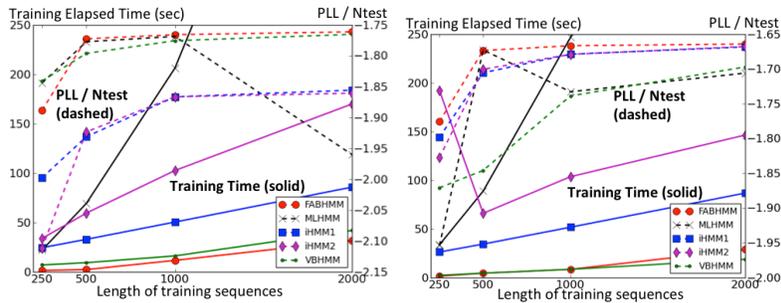

*Figure 1.* Comparison of training time and predictive log-likelihood (PLL) of FABGMM, iHMMs, VBHMM, and HLHMM (Left: Gauss, Right: Categorical).

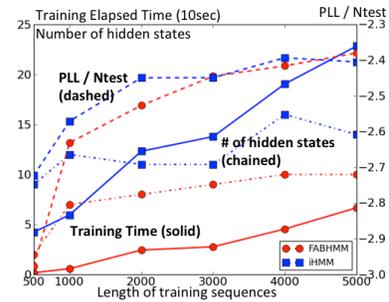

*Figure 2.* A typical behavior of FABGMM and iHMM1 (*Republic* (Plato) e-book data).

*Table 1.* The estimated number of hidden states averaged over 10 trials (Gauss/categorical). The true value was 4.

| T    | FAB     | ML      | iHMM1    | iHMM2     | VB      |
|------|---------|---------|----------|-----------|---------|
| 250  | 3.6/3.9 | 2.0/4.1 | 5.1/4.1  | 18.1/3.5  | 5.0/4.1 |
| 500  | **4.0**/4.1 | **4.0**/5.0 | 4.1/4.3 | 4.4/4.3 | 5.2/4.3 |
| 1000 | **4.0**/**4.0** | 5.7/4.7 | 4.2/4.1 | 4.3/4.1 | 6.7/4.3 |
| 2000 | **4.0**/**4.0** | 6.8/6.7 | 4.2/4.1 | 4.4/4.5 | 7.4/4.1 |
| 3000 | **4.0**/**4.0** | 7.3/7.5 | 4.2/4.2 | 4.1/4.6 | 7.8/4.2 |

Table 1 shows FABHMM almost perfectly estimated the true number of hidden states for both emission types. Surprisingly, despite FAB being an asymptotic method, it outperformed the others with relatively small data sizes. iHMMs also performed well, but their results were somehow affected by a choice of hyper-priors. VBHMM and MLHMM were significantly inferior to the others in terms of model selection performance. One plausible reason for the wrong performance of VB is the independence assumption to variables; VB approximates the marginal likelihood with ignoring the variable dependency, which makes its approximation worse than the others. BIC does not have theoretical justification in HMMs and, in fact, the MLHMM poorly performed.

Fig. 1 shows the training times and the PLLs (left: Gaussian, right: categorical). With respect to PLLs, FABHMM was competitive or slightly better than the best among the others, while none of the others did not perform well for both cases. With respect to training times, FABHMM and VBHMM were competitive and 3-4 times faster than iHMMs while all of their training time increased only linearly with the sequence length. Both PLLs and training times of MLHMM were significantly worse than those of the others. This was because MLHMM was likely to be captured by bad local minima solutions which significantly degraded both estimation performance and the convergence of its optimization while additional computational cost might mitigate this issue. An interesting observation was that FABHMM was robust for such local minima solutions because the exponentiated regularizer automatically removed such "bad" hidden states, and thus FABHMM mitigated the issue.

### 5.2. E-Book Character Sequences

Next, we evaluated the feasibility for text prediction. We prepared six books[7], *Alice's Adventures in Wonderland* (Alice), *THe Art of War* (Sunzi), *The Metamorphosis* (Kafka), *The Republic* (Plato), *The United States Declaration of Independence* (DOI), and *The Adventures of Sherlock Holmes* (Sherl). In this setting, each letter in the texts was treated as a categorical observation. The letters included some special characters, and the number of categorical alphabets varied from 32 to 50 in these data sets. The first 5000 letters of the first chapter in each book were used for training and the next 5000 letters for testing. For fair comparison, we eliminated the alphabets from the testing sets which did not appear in the training sets. Here We compared FAB with iHMM1 and VBHMM. Since these data are more complicated than artificial data, we here set $K_{\max} = 20$ for FABHMM and VBHMM.

Fig. 2 shows a typical behavior of FABHMM and iHMM and we confirmed the prediction performance of FABHMM was improved with increasing $T$. FABHMM achieved competitive prediction performance with iHMM for larger $T \geq 3000$. Although FABHMM required a longer sequence for reasonable estimation than those in the artificial simulations, we believe that the length (about 3000 - 5000) is not large in recent large scale data scenarios. Notably, the estimated number of hidden states was much smaller than that of iHMM. This means FABHMM could obtain more compact HMM representations than iHMM and that is usually desirable in practice. Table 2 shows the training times, estimated number of hidden states, and PLLs ($T = 5000$). For all data, with respect to PLLs, FABHMM and iHMM comparably performed, and VBHMM performed the worst. The training times of FABHMM were roughly three or four times faster, which agree with the results in the previous section. These results indicate clear advantages of FABHMM with respect to model selection accuracy over VBHMM, and with respect to computational efficiency over iHMM. Finally, we emphasize that FABHMM does not have hyper-parameters in its criterion, and all the above strong model selection procedures were automatically done.

---

[7]These books are available of www.gutenberg.org.

Table 2. Estimated number of hidden states $K$, training time (sec), and PLLs on the ebook data sets.

|       | FABHMM | | | iHMM | | | VBHMM | | |
|-------|---|------|------|---|-----|------|---|------|------|
| DATA  | $K$ | TIME | PLL | $K$ | TIME | PLL | $K$ | TIME | PLL |
| ALICE | 9  | 64.2 | -2.57 | 13 | 234 | -2.54 | 15 | 122.3 | -2.73 |
| KAFKA | 10 | 47.0 | -2.36 | 11 | 218 | -2.40 | 12 | 104.2 | -2.62 |
| PLATO | 10 | 66.6 | -2.38 | 14 | 228 | -2.41 | 8  | 144.9 | -2.63 |
| SHERL | 11 | 72.3 | -2.58 | 12 | 227 | -2.52 | 19 | 98.0  | -2.75 |
| SUNZI | 10 | 63.2 | -2.56 | 14 | 228 | -2.52 | 14 | 110.8 | -2.72 |
| DOI   | 10 | 73.8 | -2.98 | 12 | 232 | -2.75 | 11 | 94.6  | -2.76 |

## 6. Summary and Future Work

This paper has addressed the model selection issue for HMMs by generalizing factorized information criteria and factorized asymptotic Bayesian inference. We have theoretically shown several desirable properties (asymptotic consistency of FIC, an automatic shrinkage effect, monotonic increase in the FIC lower bound, etc) and also have experimentally shown that FABHMM outperforms the states-of-the-art variational Bayesian and non-parametric Bayesian methods with respect to model selection accuracy and computational efficiency.

We have several issues for future study. First, it is interesting to extend FAB for more flexible HMM families as iHMM has been extended for factorial HMM (Ghahramani & Jordan, 1997; van Gael et al., 2009). Second, both $FAB_{mm}$ and $FAB_{hmm}$ are designed for discrete hidden variables, and thus FAB for continuous hidden variable models is still an open problem. Third, theoretical details, such as rates of convergence (e.g., how fast the FAB iteration converges and how fast FIC converges to marginal log-likelihood), should be investigated.